\documentclass[10pt,twocolumn,letterpaper]{article}

\usepackage{iccv}
\usepackage{times}
\usepackage{epsfig}
\usepackage{graphicx}
\usepackage{amsmath}
\usepackage{amssymb}

\usepackage{pifont}
\usepackage{booktabs}

\usepackage[pagebackref=true,breaklinks=true,letterpaper=true,colorlinks,bookmarks=false]{hyperref}

\iccvfinalcopy 
\ificcvfinal\fi

\begin{document}

%%%%%%%%% TITLE
\title{Click-Feedback Retrieval}

\author{Zeyu Wang, Yu Wu\\
Princeton University \\
{\tt\small {zeyuwang,yuwu}@cs.princeton.edu}
}

\maketitle
\ificcvfinal\thispagestyle{empty}\fi

\newcommand{\zy}[1]{{\color{cyan}{(zy\@: #1)}}}

%%%%%%%%% ABSTRACT
\begin{abstract}

    Retrieving target information based on input query is of fundamental importance in many real-world applications. In practice, it is not uncommon for the initial search to fail, where additional feedback information is needed to guide the searching process. In this work, we study a setting where the feedback is provided through users clicking liked and disliked searching results. We believe this form of feedback is of great practical interests for its convenience and efficiency. To facilitate future work in this direction, we construct a new benchmark termed ``click-feedback retrieval'' based on a large-scale dataset in fashion domain. We demonstrate that incorporating click-feedback can drastically improve the retrieval performance, which validates the value of the proposed setting. We also introduce several methods to utilize click-feedback during training, and show that click-feedback-guided training can significantly enhance the retrieval quality. We hope further exploration in this direction can bring new insights on building more efficient and user-friendly search engines.
    
\end{abstract}

%%%%%%%%% BODY TEXT

\section{Introduction}

    One of the most frequent activities users perform on the Internet is searching. From learning knowledge to shopping clothes, retrieving target information by inputting a search query is always the first step. In this work, we study the issue of how to help users obtain target information effectively. Specifically, we focus on the image retrieval task for fashion product search~\cite{guo2019fashion,han2017automatic}, as it is a setting of much practical interest, and attracts lots of attention recently~\cite{kuang2019fashion,han2022fashionvil,ak2018learning,chen2020learning}. However, we note that the underlying ideas are generalizable and can be potentially applied to other searching tasks as well.

    A typical situation in practical fashion product search is that the user fails to get the target product after just a single search~\cite{kovashka2012whittlesearch}. It could be due to the user's query is ambiguous, only containing partial information of the intended product, or simply because the search engine is not strong enough and makes noisy retrieval. In such scenarios, additional information is needed to guide the search engine to retrieve the target product. Many previous works have attempted on the issue, and a popular line of works investigate the solution of utilizing extra text input as feedback~\cite{vo2019composing,chen2020learning,guo2018dialog}. Specifically, they assume that after the initial search, the user would then provide a description of the desired changes upon the retrieved product~\cite{lee2021cosmo,baldrati2022effective}. Some works have also explored other forms of feedback, for example, letting users to draw a sketch of target product~\cite{sangkloy2022sketch} or asking them questions to answer~\cite{cai2021ask}.

\begin{figure}[t]
    \centering
    \includegraphics[width=1.0\linewidth]{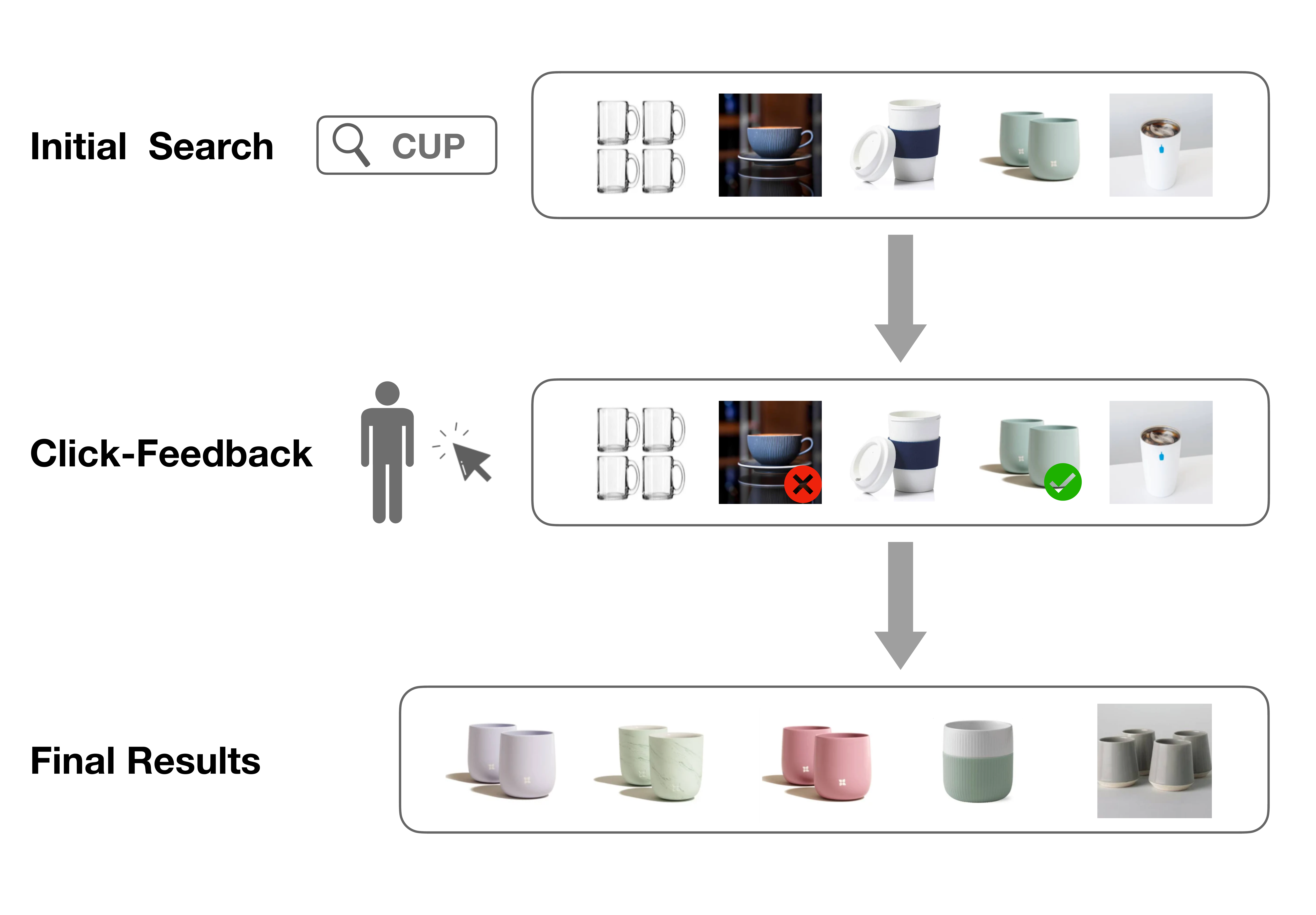}
    \caption{The diagram showing the proposed task of click-feedback retrieval. The task contains three steps. First, a text description is input as query to search target product. Then based on the initial retrieval result, a feedback agent (human in practical settings) provides feedback through clicking the liked and disliked images. Finally, the retrieval result is updated based on the initial retrieval and given click-feedback.}
    \label{fig:pull}
\end{figure}

    In this work, we instead focus on a different type of feedback, where the users only need to do a few clicks to provide their preferences. We build a new retrieval benchmark around this form of feedback and call it \textbf{click-feedback retrieval}. As Figure~\ref{fig:pull} shows, the task is composed of three steps. Initially, a text description is used as query for a first round search (the typical image retrieval setting). After the retrieval result is obtained, several top candidates are input to a feedback agent, and the agent provides feedback and returns a set of liked images (similar to target product and contain desired features) and disliked images (irrelevant or contain unwanted features). Finally, a second round of retrieval is performed based on the first-round result and received feedback. In practice, step two and three can repeat multiple times until target product is retrieved.

    In comparison to other forms, click-feedback provides several unique benefits. First and foremost is its convenience. Compared with typing extra descriptions or drawing sketches~\cite{goenka2022fashionvlp,sangkloy2022sketch}, clicking a few buttons is undeniably much more simple and efficient. In practice, this means more rounds of search can be performed within a fixed time budget. This is very beneficial in the situation where the user does not have the exact target features in mind before the search (\eg looking for a cup but does not have other details otherwise) and is forming the preference through browsing along the searching process. More rounds of search exposes the user with more candidates and thus better helps the user form the preference. Besides, click-feedback is also helpful in the case where the desirable feature is hard to describe in language, \eg a specific shape or texture that is uncommon.

    To facilitate future work on studying how to better incorporate click-feedback into retrieval, we construct a new benchmark based on Fashion200K dataset~\cite{han2017automatic}. One challenge of building the benchmark is that the feedback needs to be generated dynamically online based on the current retrieval result, but it is not easy to have human-in-the-loop training in reality. We tackle the issue by approximating the human preference with a strong image encoder~\cite{han2022fashionvil} and find it work reasonably well in practice. We experiment with several methods that can utilize click-feedback to update retrieval, and the result shows that the retrieval performance can be improved dramatically after incorporating the click-feedback. This validates the effectiveness of the proposed setting. 

    As a summary, we make the following contributions in this work:

        \begin{itemize}
            \item We study a previously less-explored form of feedback in the fashion image retrieval setting, where the feedback is provided through users clicking groups of liked and disliked images after the initial search.

            \item We introduce a new task named click-feedback retrieval and construct a benchmark to facilitate future work in this direction. 

            \item We experiment with a training-free method to incorporate click-feedback in retrieval and demonstrate significant improvement of retrieval performance (R@10 being improved from 41.7\% to 51.1\%, and median rank being halved from 18 to 9), which shows the effectiveness of the proposed setting.

            \item We further propose methods to train the model with click-feedback, and show additional enhancement of performance over inference-only baseline (R@10 being improved from 51.1\% to 58.5\%, and median rank being reduced from 9 to 5).

        \end{itemize}

\section{Related Work}

\paragraph{Text-image retrieval.}
Text-image retrieval has been extensively studied by many researchers due to its high real-world application value. The scenario is to retrieve images of one modality with a given query text of another modality. Existing methods calculate the similarity of each text-image pair by mapping the input of the two modalities to the same feature space. 
To extract text and image features, early works~\cite{fromedeep,chun2021probabilistic,hu2021learning,faghri2017vse++,zheng2020dual} mainly focus on visual semantic embedding with regard to data and the loss function respectively which provides high-efficiency baselines. Further, \cite{lee2018stacked,chen2020uniter,cui2021rosita,qu2021dynamic} leverage cross-attention and self-adaptive approaches to explore the interaction between the text and image data deeply. 
After the feature extraction stage, some works propose aligning cross-modal features for better representations. \cite{faghri2017vse++,wang2018learning,zheng2020dual} pay attention to global alignment while \cite{lee2018stacked,wang2019camp} follow interest with local alignment.
Beyond the above, some works raise the retrieval efficiency by hash encoding~\cite{yang2017pairwise,zhang2018attention} and model compression~\cite{gan2022playing,huang2004aligning}.
Recently, many researchers~\cite{lu2019vilbert,kim2021vilt,li2021align} have begun to design different model architectures, which promote retrieval performance by a large margin. Some~\cite{lu2019vilbert,kim2021vilt,li2021align} design pre-training pretext tasks to obtain more discriminate features in an end-to-end manner. Others~\cite{yao2021filip,li2021align} concentrate on increasing the scale of pre-training data which naturally boosts the downstream retrieval task.

% Text-image retrieval has been extensively studied due to its practical value in many real-world applications.The setting is to retrieve images of interests given text descriptions as query. To accomplish the task, existing methods calculate the similarity between text and image pairs by mapping their inputs to the same feature space, where a simple metric like cosine distance is used for calculating the similarity. Early works~\cite{fromedeep,chun2021probabilistic,hu2021learning,faghri2017vse++,zheng2020dual} 

% it is desirable to align cross-modal features to compute pairwise similarity and achieve retrieval. 

\paragraph{Retrieval with feedback.} Since the correspondence between text and image is full of diversity and uncertainty, it is often difficult to obtain target image at one shot. Often times, addition feedback information is needed to adjust the retrieval results. To this end, many works have studied a variety of feedback methods, including using absolute attributes~\cite{zhao2017memory,han2017automatic,ak2018learning}, relative attributes~\cite{parikh2011relative,kovashka2012whittlesearch,yu2019thinking}, attribute-like modification text~\cite{vo2019composing}, and natural language~\cite{guo2018dialog,guo2019fashion}. Other works also explored on using sketches or asking questions~\cite{sangkloy2022sketch,cai2021ask}. In this work, we study a different type of feedback, which we call click-feedback, where the users provide feedback through clicking the liked and disliked images. It provides much convenience and efficiency compared to other feedback forms. Click-feedback retrieval resembles and draws inspirations from a classic line of works on relevance feedback~\cite{rui1998relevance,zhou2003relevance,ruthven2003survey}. However, we note that early works on it are done in the pre-deep learning era and there haven't been much focus on it recently for image retrieval with deep neural networks~\cite{putzu2020convolutional}.

%Because text is the most natural way of interaction between humans and models, this way of using text feedback to convey concrete information to search engines can effectively help users retrieve the desired images.

\begin{figure*}[t]
    \centering
    \includegraphics[width=1.0\linewidth]{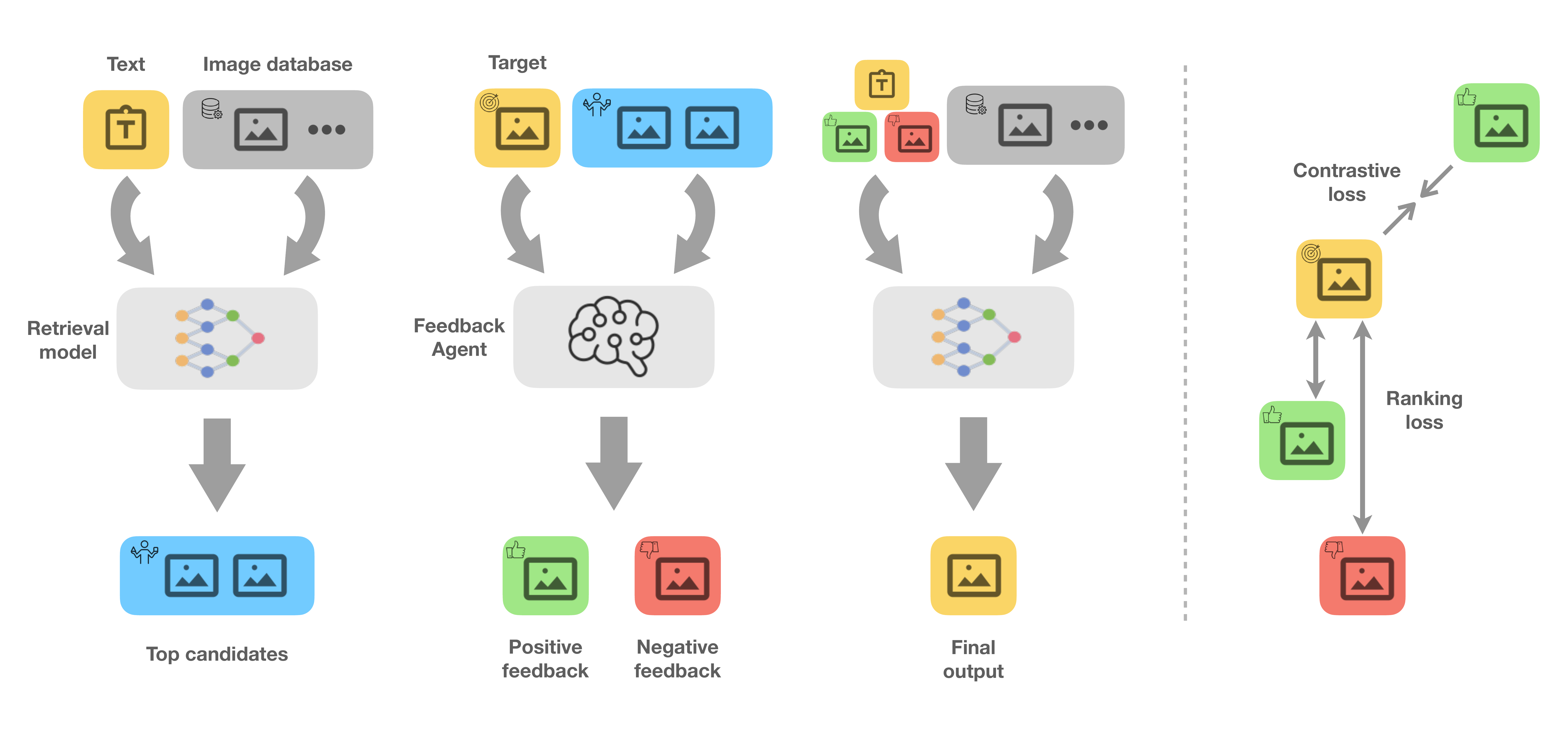}
    \caption{\textit{Left}: concrete implementation of the proposed click-feedback retrieval process. It consists of three steps. First, given input text description, the retrieval model searches the image database and generates initial retrieval. Second, top candidates from first step is input to feedback agent, which, based on the similarity to the target, outputs a set of positive feedback images and a set of negative feedback images (simulating user clicking likes and dislikes). Finally, the retrieval model outputs the final retrieval result based on the text and the click-feedback generated in the second step. \textit{Right}: illustration of two type of losses for click-feedback-guided training.}
    \label{fig:method}
\end{figure*}

\section{Click-feedback Retrieval}
\label{sec:setting}

    Retrieving target information is a fundamental operation people interact with Internet. In this section, we formalize this interaction and introduce our proposed setting of retrieval with click-feedback. We will focus on the scenario of product search, where a user inputs a text description of the target product as query and the search engine returns a list of candidates  in the form of images. But note that the underlying idea can be generalized to other scenarios with potentially different input-output format.

    Given a user text query $q$, the aim of search engine is to retrieve target image $i^t$ from a large group of all candidates $\mathbb{G}^{all}$, where $\mathbb{G}^{all} \doteq \{i_1, i_2, \dots, i_n\}$, and $i^t \in \mathbb{G}^{all}$. Essentially, the retrieval operation performs a ranking of elements in $\mathbb{G}^{all}$ based on $q$, \ie it gives a rank for each image, $r_q(i)$, such that $r_q(i) \in \{1,2,\dots,n\}$ and $\{r_q(i_1), r_q(i_2), \dots, r_q(i_n) \} = \{1,2,\dots,n\}$ (lower rank means better alignment with $q$). Under this notation, a successful retrieval would have $r_q(i^t)$ as small as possible. 

    In practice, unsatisfactory searches are very common, where $r_q(i^t)$ is large (target product is not contained in the first few returned pages). This could be due to various factors, and a frequent one is that the input query $q$ is not specific enough such that too many candidates can be matched to it. Under such situation, additional information is needed to help retrieve the target product. Therefore previous works have tackled on various ways to provide the necessary information through feedback, \eg by adding additional language descriptions, sketches of target product, or asking questions~\cite{lee2021cosmo,sangkloy2022sketch,cai2021ask}. In this work, we argue for a previously less explored setting where the feedback is in the form of clicking the likes and dislikes. The main benefit of it is convenience, as a simple click is much easier than typing sentences or drawing pictures.  

    Specifically speaking, after the initial retrieval, users can view the top-k retrieved products, and then select among them the ones $\mathbb{G}^{like}$ they like (containing desirable features and want to see more), and the ones $\mathbb{G}^{dislike}$ they dislike (containing undesirable features and want to see less). Formally, we define feedback $f \doteq \{ \mathbb{G}^{like}, \mathbb{G}^{dislike} \}$. And the updated retrieval would generate a new ranking $r_{q,f}$ and the aim is to improve the search result such that $r_{q,f}(i^t) \ll r_q(i^t)$ . The complete three-step process of the proposed click-feedback retrieval is summarized in the left part of Figure~\ref{fig:method}.

    \paragraph{Evaluation.} We adopt the widely-used evaluation metrics in retrieval community~\cite{wang2022multi,han2022fashionvil}, \ie R@K (recall at rank K, higher the better), median and mean rank (lower the better). Formally, R@K is defined as the fraction of test instances where $r(i^t) < K$. Following previous works, we report R@1, R@5 and R@10. Median and mean rank are median and mean of $r(i^t)$ among all test instances respectively.

\section{Methods}
\label{sec:method}

    In this section, we propose several methods that tackle the setting of click-feedback retrieval introduced in Section~\ref{sec:setting}. They will serve as baselines for future works in this direction. Broadly, they can be divided into two categories, one without training and the other with training.

    \subsection{Training-free inference}

        Given an input query $q$, the language encoder $E_l$ embeds $q$ to a vector $v_q$, and correspondingly the vision encoder $E_v$ embeds image $i$ to a vector $v_i$ in the same latent space, \ie $v_q, v_i \in \mathbb{R}^d$. Then the retrieval rank of the image, $r(i)$, is generated based on some measure of similarity $\mathbf{S}$ between $v_i$ and $v_q$. Usually, the cosine similarity is used for its simplicity. Therefore, the ranking function for the setting without feedback is:

            \begin{equation}
                \mathcal{R}^{NF} = \mathbf{S}(v_i, v_q)
            \end{equation}

        When click-feedback is available, the ranking function can be updated with,

            \begin{equation}
                \label{eq:feedback}
                \begin{split}
                    \mathcal{R}^{F} = \mathbf{S}(v_i, v_q) + \lambda_p \mathbf{S}(v_i, \mathbb{G}^{like}) \\
                            - \lambda_n \mathbf{S}(v_i, \mathbb{G}^{dislike})
                \end{split}
            \end{equation}

        where the similarity between an image $i$ with a group of images $\mathbb{G}$ can be defined as the average similarity between $i$ and images in $\mathbb{G}$:

            \begin{equation}
                \mathbf{S}(v_i, \mathbb{G}) = \frac{1}{|\mathbb{G}|}~\sum_{i' \in \mathbb{G}}
                    ~\mathbf{S}(v_i, v_{i'}) 
            \end{equation}

        Intuitively, $\mathcal{R}^{F}$ up-weights a candidate image by its similarity to the images liked by the user, and down-weights with the similarity to those disliked. And the coefficients $\lambda_p$ and $\lambda_n$ control the relative contribution of the positives and negatives.

    \subsection{Training methods}
    
        The previous section introduces how to adapt an existing model to incorporating click-feedback during inference. When the feedback is available during model development, additional training techniques can be utilized to further improve the performance. Specifically, we experiment with two different loss functions.  

        \paragraph{Ranking loss.} For ranking loss, we encourage the similarity between the target image and the positive feedback images (liked ones) to be larger than the similarity between the target image and the negative feedback images (disliked ones),

            \begin{equation}
            \label{eq:ranking}
                \begin{split}
                    \mathcal{L}_{feedback}^{r} = 
                    \max ( 0, -\mathbf{S}(v_i, \mathbb{G}^{like}) \\
                            +~\mathbf{S}(v_i, \mathbb{G}^{dislike}) +~m )
                \end{split}
            \end{equation}
        
        where $m$ is a hyperparameter to control the margin of the separation. 

        \paragraph{Contrastive loss.} For contrastive loss, we encourage the distance between embeddings of the target image and matched feedback images to be small, and the distance between embeddings of the target image and mismatched feedback images to be large. Only positive feedback images are used in the contrastive loss here, as empirically we find contrasting away negative feedback images hurts the learned representation. Concretely, the loss is defined as:

            \begin{align}
            \label{eq:constrastive}
                &\mathcal{L}_{i^t2{i^f}} = -\frac{1}{B}\sum_j^B \log~\frac{\exp{(E_v(i^t_j)^TE_v(i^f_j)/t)}}{\sum_{k=1}^B \exp{(E_v(i^t_j)^TE_v(i^f_k)/t)}} \notag \\
                &\mathcal{L}_{i^f2i^t} = -\frac{1}{B}\sum_j^B \log~\frac{\exp{(E_v(i^f_j)^TE_v(i^t_j)/t)}}{\sum_{k=1}^B \exp{(E_v(i^f_j)^TE_v(i^t_k)/t)}} \notag \\
                &\mathcal{L}_{feedback}^{c} = \frac{1}{2}(\mathcal{L}_{i^t2{i^f}} + \mathcal{L}_{i^f2i^t}) 
            \end{align}

        where $i^t$ is the target image, and $i^f$ is the positive feedback image.
        
        \paragraph{Text-image alignment.} As the feedback losses mentioned before only updates the image encoder, to avoid the learned image representation deviating too much from the text representation, a text-image alignment loss is also added during feedback training to keep image and text embedding aligned. Specifically, we use a constrastive loss similar to equation \ref{eq:constrastive}:

            \begin{align}
            \label{eq:retrieval}
                &\mathcal{L}_{t2i} = -\frac{1}{B}\sum_j^B \log~\frac{\exp{(E_l(q_j)^TE_v(i_j)/t)}}{\sum_{k=1}^B \exp{(E_l(q_j)^TE_v(i_k)/t)}} \notag \\
                &\mathcal{L}_{i2t} = -\frac{1}{B}\sum_j^B \log~\frac{\exp{(E_v(i_j)^TE_l(q_j)/t)}}{\sum_{k=1}^B \exp{(E_v(i_j)^TE_l(q_k)/t)}}  \\
                &\mathcal{L}_{ti-align} = \frac{1}{2}(\mathcal{L}_{t2i} + \mathcal{L}_{i2t}) \notag
            \end{align}

        % where $i$ is the product image and $q$ is the input text query.

        The total loss for training with click-feedback is then:

        \begin{equation}
        \label{eq:feedback_all}
            \mathcal{L}_{all} = \mathcal{L}_{feedback} + \mathcal{L}_{ti-align}
        \end{equation}
        
        After training, equation~\ref{eq:feedback} is used as the final ranking function during inference as before.

\section{Experiments}

    In this section, we first introduce the concrete setup of a benchmark for retrieval with click-feedback, including the dataset and how click-feedback is generated. Then we elaborate on the implementation details on model architecture and training. Finally, experiment results are shown with detailed analysis on the effectiveness of the proposed setting.

    \subsection{Experimental setup}

\begin{table*}[t]
    \begin{center}
        \begin{tabular}{lccccccc}
     \toprule
       \textbf{Method}  & \textbf{Feedback} & \textbf{Training} & \textbf{R@1} $\uparrow$ & \textbf{R@5} $\uparrow$ & \textbf{R@10} $\uparrow$ &  \textbf{MedR} $\downarrow$ & \textbf{MeanR} $\downarrow$ \\
    
    \midrule
      Baseline   & \ding{55} & \ding{55} & 13.8  & 31.7 & 41.7 & 18 & 173.0 \\

    \midrule

      Inference-only & \ding{51} & \ding{55} & 41.7 & 46.3 & 51.1 & 9 & 155.2 \\

      Contrastive loss & \ding{51} & \ding{51} & 41.1 & 47.7 & 53.8 & 7 & 104.5 \\

      Contrastive loss + SepEnc & \ding{51} & \ding{51} & 41.1 & 47.6 & 55.2 & 7 & 79.9 \\

      Ranking loss & \ding{51} & \ding{51} & 43.1 & 48.8 & 54.2 & 6 & 95.7 \\

      Ranking loss + SepEnc & \ding{51} & \ding{51} & 39.1 & 50.6 & \textbf{58.5} & \textbf{5} & \textbf{70.5} \\

    \bottomrule
    \end{tabular}
    \end{center}
    
    \caption{Performance of different methods on Fashion200K~\cite{han2017automatic} retrieval task. \textit{Baseline} is a CLIP~\cite{radford2021learning} model finetuned on the dataset without feedback. \textit{Inference-only} is the \textit{baseline} model with click-feedback added during test.\textit{Contrastive loss} and \textit{Ranking loss} use click-feedback as additional supervision during training. \textit{Contrastive loss+SepEnc} and \textit{Ranking loss+SepEnc} use two separate image encoders, one for computing cross-modal text-to-image similarity, and the other one for computing unimodal image-to-image similarity.} 
    \label{tab:main_res}
\end{table*}

        \paragraph{Benchmark.} We build our retrieval with click-feedback benchmark upon the Fashion200K dataset~\cite{han2017automatic}, which is a large-scale dataset containing more than 200,000 clothing images spanning across five major fashion categories (dress, top, pants, skirt and jacket) with various styles. The dataset comes with different types of annotations including detailed product information and bounding boxes. We only use the images and corresponding attribute-based text descriptions (\eg ``black roll-up sleeve blouse'') for our experiments.

        \paragraph{Click-feedback. } Ideally the feedback of likes and dislikes should be provided by human to simulate the real use case. However, in practice it is hard to train models with human in-the-loop, especially considering the training process can easily contain hundreds thousands of iterations. Therefore, to make training feasible, we need other ways to simulate the feedback process automatically without human involvement. This boils down to generating the similarities between the candidate images and the target image. One way to obtain this is to utilize a good image encoder network and compute the similarities in its latent space. Another way is to approximate the image similarities with the similarities of the corresponding text annotations (\eg calculating the intersection-over-union of the ground-truth attributes). In this work, we utilize the former approach as we find empirically that the generated similarities are more fine-grained using the dense representation from an image encoder. 

        \paragraph{Implementation details.} To approximate human feedback, we use the FashionViL model proposed by Han \etal ~\cite{han2022fashionvil}. FashionViL is a fashion-focused vision-language model with specific designs that fully exploit the specialties in fashion domain. We utilize the model released by the authors that was pretrained with over 1.35 million image-text pairs from several public fashion-related datasets, including Fashion200K, the dataset we build our benchmark on. The model is only used to simulate human preference and is not modified or used for retrieval in the experiments. Concretely, after the initial retrieval, FashionViL model computes the similarities between the top ten retrieved images and the target image, and outputs the most similar one to target image as $\mathbb{G}^{like}$ and the least similar one as $\mathbb{G}^{dislike}$.

        For the retrieval model, we utilize CLIP~\cite{radford2021learning}, which has been used as initialization for many vision-language tasks recently due to its great transferability. Since CLIP is not designed for fashion product retrieval, we first finetune it on Fashion200K dataset to have a better starting point (avoiding the situation where no relevant images are among top ten after the initial retrieval for a reasonable feedback). Specifically, we use the publicly available `ViT-B/32' model and finetune it on Fashion200K training set for 30 epochs with the loss of equation~\ref{eq:retrieval}. AdamW optimizer is used with a cosine learning rate scheduler with max learning rate of 3e-6 and a linear warm up of 5 epochs. After finetuning, the median rank (lower the better) on test set decreases from 135 to 18. 
        
        With this as initial point, we train the model with click-feedback for another 30 epochs using the feedback loss of equation~\ref{eq:feedback_all}. The margin $m$ is set to 0.2 for ranking loss and the same optimizer and scheduler configuration is used. During test, we use equation~\ref{eq:feedback} to rank all candidate images based on the input text description as well as the feedback given after the initial retrieval. We set $\lambda_p$ as 1.0 and $\lambda_n$ as 0.5 to give a higher weight to positive feedback.

    \subsection{Results without training}

        The \textit{inference-only} entry in Table~\ref{tab:main_res} shows the result when adding additional feedback of liked and disliked images only during inference. Compared to \textit{baseline} (the initial retrieval result without feedback), there is a dramatic enhancement in performance. The median rank is halved from 18 to 9, and the mean rank is decreased from 173.0 to 155.2. The R@10 is improved by an absolute of 9.3\% (from 41.7\% to 51.0\%). Note that there is even larger improvement of R@1 and R@5 (from 13.8\% to 41.7\% for R@1 and from 31.7\% to 46.3\% for R@5). However, that increase is mainly contributed by the instances where the target image is retrieved as top 10 during the initial retrieval, resulting the positive feedback image to be the target image itself. So we will mostly focus on the improvement of R@10, median rank and mean rank, and include R@1 and R@5 here for completeness. 

        \paragraph{Influence of positive and negative feedback.} Table~\ref{tab:pos_vs_neg} shows how positive and negative feedback contributes to the overall improvement by varying the coefficients $\lambda_p$ and $\lambda_n$ in equation~\ref{eq:feedback}. The second row ($\lambda_p=1.0$, $\lambda_n=0.0$) shows the result when only liked images are used as feedback to update retrieval. As can be seen, the positive feedback accounts for most of the improvement. R@10 is improved from 41.7\% to 49.4\%, and median rank is improved from 18 to 11,  compared to R@10 of 51.1\% and median rank of 9 when using full feedback. The third row ($\lambda_p=0.0$, $\lambda_n=0.1$) shows the result with only disliked images used as feedback \footnote{Note that here a smaller number for $\lambda_n$ is used to further down-weight the contribution of negative feedback in equation~\ref{eq:feedback}, as otherwise the totally irrelevant images would be given a high score (as they are most dissimilar to the negative feedback images) and dominate the retrieval (R@10 is 2.7\% and median rank is 1005 when using $\lambda_p=0.0$, $\lambda_n=1.0$).}. It manages to introduce improvement over baseline, improving R@10 from 41.7\% to 43.6\% and meidan rank from 18 to 16. But the enhancement of performance is relative small compared to using only positive feedback. This shows the positive examples are relatively more effective in helping the retrieval. Intuitively, this is because positive feedback provides a more direct guidance. Despite this, the negative feedback is still useful as the improvement it introduces is complementary to that of positive feedback, as can be seen from the last row of Table~\ref{tab:pos_vs_neg}.

\begin{table}[t]
    \begin{center}
        \begin{tabular}{ccccc}
             \toprule
               $\lambda_p$ & $\lambda_n$ & \textbf{R@10} $\uparrow$ &  \textbf{MedR} $\downarrow$ & \textbf{MeanR} $\downarrow$ \\
            \midrule
            0.0 & 0.0 & 41.7 & 18 & 173.0 \\
            \midrule
            1.0 & 0.0 & 49.4 & 11 & 148.6 \\
            0.0 & 0.1 & 43.6 & 16 & 170.5 \\
            1.0 & 0.5 & 51.1 & 9 & 155.2 \\
            \bottomrule
        \end{tabular}
    \end{center}
    
    \caption{Influence of positive and negative feedback for retrieval performance. $\lambda_p$ and $\lambda_n$ are coefficients in equation~\ref{eq:feedback}. Positive feedback introduces more improvement over negative feedback, but the two are complementary to each other and give the best performance when combined together.}
    \label{tab:pos_vs_neg}
\end{table}

    \subsection{Results with training}

        While using feedback only during inference has already introduced much improvement, additional increase in performance is achieved using feedback-based training. As shown in Table~\ref{tab:main_res}, feedback-guided training with either contrastive loss or ranking loss can boost the performance. Concretely, \textit{Contrastive loss} helps increasing R@10 from 51.1\% to 53.8\%, improving median rank from 9 to 7, and mean rank from 155.2 to 104.5. \textit{Ranking loss} provides even larger improvement, which increases R@10 by 3.1\% (from 51.1\% to 54.2 \%), and reduces median rank and mean rank from 9 to 6 and from 155.2 to 95.7 respectively. This validates the effectiveness of the proposed training with click-feedback. We assume the reason why \textit{Ranking loss} works better than \textit{Contrastive loss} is that it utilizes both positive and negative feedback images, while \textit{Contrastive loss} only utilizes the positive ones (we experimented on using negative feedback with contrastive loss as well but that fails to introduce improvement, as contrasting away negative feedback tends to hurt the learned representation). This again shows the unique value provided by the negative feedback, as it provides complementary information to the positive feedback.
        
        Empirically, we find that further improvement can be obtained by using separate image encoders for calculating cross-modal text-to-image similarity (the $\mathbf{S}(v_i, v_q)$ in equation~\ref{eq:feedback}) and unimodal image-to-image similarity (the $\mathbf{S}(v_i, \mathbb{G}^{like})$ and $\mathbf{S}(v_i, \mathbb{G}^{dislike})$ in equation~\ref{eq:feedback}). Concretely, \textit{Contrastive loss+SepEnc} improves over \textit{Contrastive loss} on R@10 for 1.4 points (from 53.8\% to 55.2\%) and largely reduces mean rank (from 104.5 to 79.9). Similarly, \textit{Ranking loss+SepEnc}
        enhances R@10 by an additional 4.3 points (from 54.2\% to 58.5\%), and reduces median rank to 5 and mean rank to 70.5 (from 95.7). 
        
        We assume the reason why using separate encoders helps might be that there is some inherent differences between multi-modal embedding space and uni-modal embedding space, which is hard to reconcile within one model (at least for the model used in our experiment). Therefore, using separate encoders avoids the interference when trying to capture both text-to-image similarity and image-to-image similarity. And the reason why \textit{Ranking loss} enjoys more improvement than \textit{Contrastive loss} after using separate encoders could be that it provides separation of two different type of loss (ranking and contrastive), as the text-to-image alignment uses contrastive loss as well. However, note that using separate encoders comes at the cost of increased parameters and computation cost. We leave to the future work for addressing the issue.

    \subsection{Additional experiments}
    \label{sec:more_exp}

        \paragraph{Number of feedback instances.} For all the experiments shown previously, the feedback agent provides only one positive image and one negative image. Here we change that assumption and vary the number of images provided as feedback. The result is shown in Table~\ref{tab:feedback_num}, where $n_{like}$ is the number of positive feedback images in $\mathbb{G}^{like}$ and $n_{dislike}$ is the number of negative feedback images in $\mathbb{G}^{dislike}$. From first five rows of the table (number of positive/negative feedback images increasing from one to five), increasing the number of feedback instances doesn't help in this case. While the performance on mean rank increases, the performance of R@10 and median rank drops. The last two rows show that the performance drop of R@10 and median rank mainly comes from the increasing of positive feedback instances. We find that this is because the false positives, as simply choosing top five most similar images as positive feedback could include images that are in fact not similar to the target image. Therefore, we keep $n_{like}$ and ${n_{dislike}}$ to be one in all other experiments.

\begin{table}[t]
    \begin{center}
        \begin{tabular}{ccccc}
             \toprule
               $n_{like}$ & $n_{dislike}$ & \textbf{R@10} $\uparrow$ &  \textbf{MedR} $\downarrow$ & \textbf{MeanR} $\downarrow$ \\
            \midrule
            1 & 1 & 51.1 & 9 & 155.2 \\
            \midrule
            2 & 2 & 51.1 & 10 & 137.2 \\
            3 & 3 & 50.3 & 10 & 134.3 \\
            4 & 4 & 49.3 & 11 & 135.3 \\
            5 & 5 & 48.3 & 12 & 139.1 \\
            5 & 1 & 47.5 & 13 & 146.9 \\
            1 & 5 & 51.5 & 9 & 150.3 \\
            \bottomrule
        \end{tabular}
    \end{center}

    \caption{Performance of \textit{Inference-only} model with varying number of feedback instances.}

    \label{tab:feedback_num}
\end{table}

        \paragraph{Adding diversity for feedback candidates.} In previous experiments, the top ten images based on the similarity to input text description are given to the feedback agent as candidates for generating the feedback $\mathbb{G}^{like}$ and $\mathbb{G}^{dislike}$. However, it is not required to only use this criterion, and choosing which set of images to ask for feedback is a design choice that can be changed. Here we experiment with a heuristic that tries to increase the visual diversity of the candidate images. The intuition is that this could avoid the situation where the top retrieved images are too similar to each other and the feedback based on them doesn't provide enough information. Specifically, we utilize an iterative method to select candidate images ($\{i^c\}$):

            \begin{equation}
            \label{eq:diversity}
             \begin{split}
                 i_n^c =~&arg max_i~(\mathbf{S}(v_i, v_q) \\
                        &-~\lambda_{diversity}\mathbf{S}(v_i, \{i_1^c, i_2^c, \dots, i_{n-1}^c\}))  
             \end{split}
            \end{equation}

\begin{table}[t]
    \begin{center}
        \begin{tabular}{cccc}
             \toprule
               $\lambda_{diversity}$ & \textbf{R@10} $\uparrow$ &  \textbf{MedR} $\downarrow$ & \textbf{MeanR} $\downarrow$ \\
            \midrule
            0.0  & 51.1 & 9 & 155.2 \\
            \midrule
            0.2 & 51.2 & 9 & 151.5 \\
            0.4 & 50.9 & 9 & 151.6 \\
            0.6 & 50.0 & 10 & 153.9 \\
            0.8 & 48.8 & 12 & 151.3 \\
            1.0 & 47.1 & 13 & 155.8 \\
            \bottomrule
        \end{tabular}
    \end{center}

\caption{Performance of \textit{Inference-only} model with a diversity heuristic to select visually-different images as candidates to receive feedback. }
    
    \label{tab:diversity}
\end{table}

\begin{figure*}[t]
    \centering
    \includegraphics[width=1.0\linewidth]{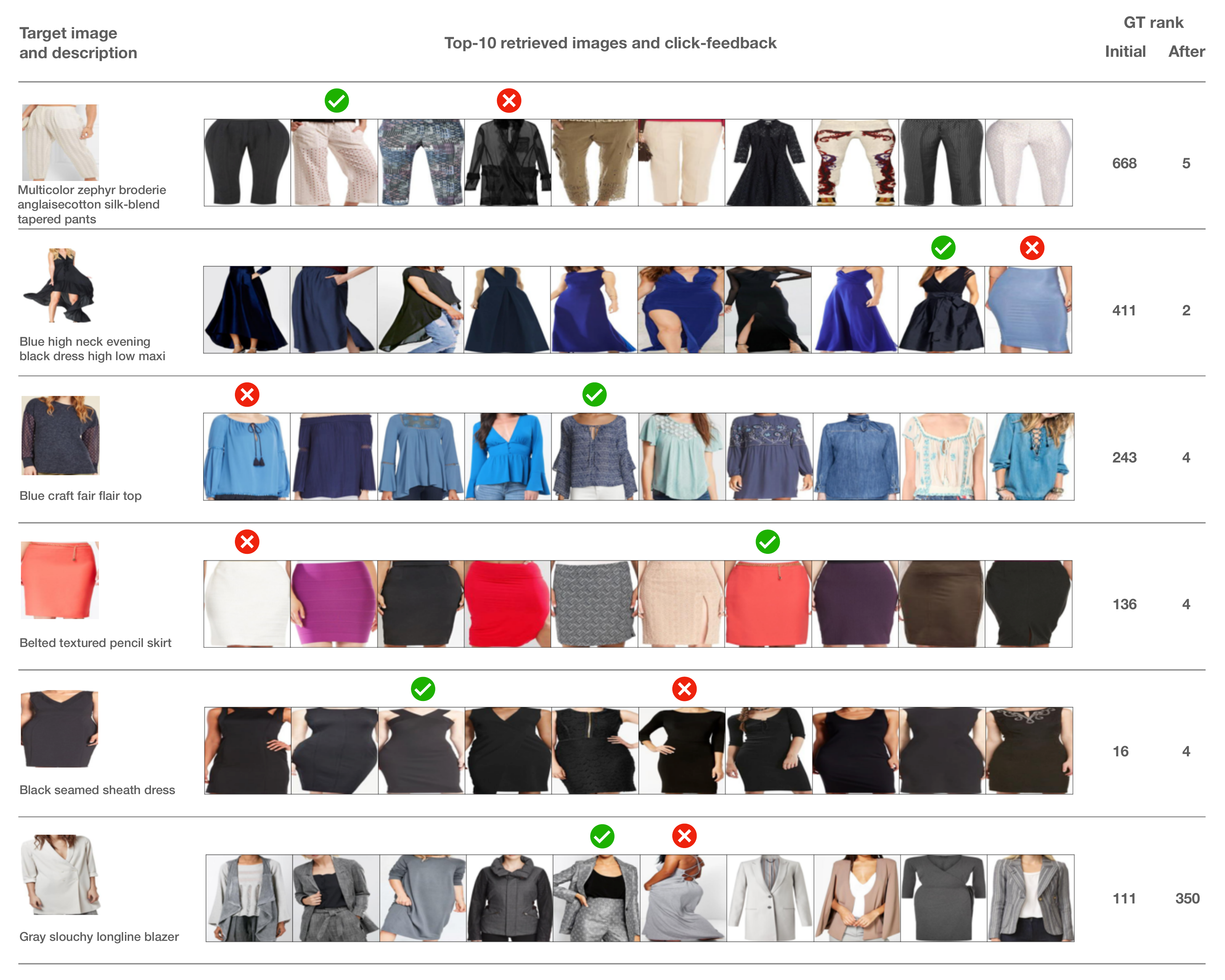}
    \caption{Qualitative examples. First five rows show how click-feedback helps retrieving target image to top-ten by utilizing the rich visual information contained in it. Notice how it drastically improves the performance when the target is ranked way back in the initial retrieval (first few rows). The last row shows a typical failure case where the performance degrades with feedback. It can be attributed to poor initial retrieval, where the feedback agent provides a dissimilar image as positive feedback.}
    \label{fig:qualitative}
\end{figure*}

        which down-weights the images which are too similar to those already selected, and $\lambda_{diversity}$ here controls the degree of the diversity. Table~\ref{tab:diversity} shows the results with $\lambda_{diversity}$ ranging from 0.2 to 1. Here the added diversity doesn't help on the performance. We find that this is because for Fashion200k and the CLIP model used, the candidate set through naive selection of top ten images after initial retrieval is already very diverse. The additional diversity heuristic described by equation~\ref{eq:diversity} doesn't add more on the diversity side but decreases the quality of the candidates. However, we note that this might help in practice as the web is full of very similar and often identical products.

\section{Limitations and future work}

    We believe the task of retrieval with click-feedback holds great promise on improving the efficiency of retrieval and the overall user satisfaction in real-world use cases. We have demonstrated such penitential through various experiments introduced in previous sections. Here we list several promising directions to explore in the future.

    In this work, we only focus on one round of feedback for simplicity. It is natural to extend the task to retrieval with multiple rounds of click-feedback. This better approximates the real-world searching scenarios and introduces the interesting challenge of how to handle the long history of interactions, which is not covered in the single-round feedback setting studied in this paper. 

    For feedback-guided training, it would be interesting to explore how to utilize reinforcement learning to directly supervise the retrieval model using the final groundtruth rank as reward. Concretely, the reword would encourage the final groundtruth rank after the feedback to be as small as possible. This is a more intuitive supervision as that directly optimizes the real target of the retrieval. Besides, it brings another benefit of supervising the three steps (initial retrieval, feedback generation, retrieval update) as a whole. In theory, this could lead to a better policy on choosing which set of candidates for receiving feedback (instead of always choosing the top retrieved instances after initial retrieval or using some manually-designed heuristics as we explored in section~\ref{sec:more_exp}). However, we note that it is not straightforward on how to properly train with reinforcement learning to supervise the whole three-step click-feedback retrieval process using the final retrieval rank as reward. We leave to future work for advancing in this direction.

    Finally, to avoid human-in-the-loop training, we utilized a strong image encoder to approximate the human preference. While we found it work well in practice, it could be imperfect at times and provide incorrect feedback. Furthermore, it is not easy to quantitatively evaluate how much noise it introduces. Therefore, it would be of interests to explore other methods to generate click-feedback that better capture human judgement.

\section{Conclusion}

    As a summary, in this work we study how to help users retrieving target information efficiently during search. In this regard, we focus on a previously less-explored setting where the user provides feedback through clicking a set of liked and disliked images after seeing the initial retrieval results. We proposed a new task termed \textbf{click-feedback retrieval} and built a large-scale benchmark in the fashion product retrieval domain around it. We introduced several methods to incorporate click-feedback and demonstrated that the retrieval performance can be improved significantly. We believe further efforts on the task would greatly benefit the field and help building more efficient and user-friendly search engines for real-world applications.

\clearpage

{\small
\bibliographystyle{ieee_fullname}
\bibliography{main}
}

\end{document}